\documentclass[sigconf]{acmart}

\copyrightyear{2026}
\acmYear{2026}
\setcopyright{cc}
\setcctype{by}
\acmConference[CIKM '26]{Proceedings of the 35th ACM International Conference on Information and Knowledge Management}{November 07--11, 2026}{Rome, Italy}
\acmBooktitle{Proceedings of the 35th ACM International Conference on Information and Knowledge Management (CIKM '26), November 07--11, 2026, Rome, Italy}
\acmDOI{10.1145/3799682.3841092}
\acmISBN{979-8-4007-2539-5/2026/11}

\AtBeginDocument{%
  }

\makeatletter
\let\ACM@origbaselinestretch\baselinestretch
\makeatother

\usepackage{booktabs}
\usepackage{graphicx}
\usepackage{colortbl}
\usepackage{multirow}
\usepackage{array}
\usepackage{tabularx,booktabs}
\newcolumntype{Y}{>{\centering\arraybackslash}X}
\usepackage{enumitem}
\newcommand{\proposed}{\textsf{CoSPOT}}
\usepackage{xcolor}
\usepackage{pifont}
\usepackage{amssymb}
\usepackage{bbding}
\usepackage{xcolor}
\usepackage{algorithm}
\usepackage{algpseudocode}
\usepackage{amsmath}
\usepackage{amssymb}

\usepackage{makecell}
\usepackage{cuted}
\usepackage{caption}

\begin{document}

\title{Compositional Spectral Prompts for LLM-based Online Time Series Forecasting}

\author{Seungyoon Choi}
\affiliation{%
  \institution{KAIST}
  \city{Daejeon}
  \country{Republic of Korea}}
\email{csyoon08@kaist.ac.kr}

\author{Hyunchul Kim}
\affiliation{%
  \institution{KAIST}
  \city{Daejeon}
  \country{Republic of Korea}}
\email{khchul@kaist.ac.kr}

\author{Jae-Gil Lee}
\affiliation{%
  \institution{KAIST}
  \city{Daejeon}
  \country{Republic of Korea}}
\email{jaegil@kaist.ac.kr}

\author{Chanyoung Park}
\affiliation{%
  \institution{KAIST}
  \city{Daejeon}
  \country{Republic of Korea}}
\email{cy.park@kaist.ac.kr}
\renewcommand{\shortauthors}{Choi et al.}

\begin{abstract}
  To address the sequential and evolving nature of time series, the Online Time Series Forecasting (OTSF) task has been extensively studied in multiple domains. Existing research focuses on adapting to non-stationary environments by employing memory buffer-based retrieval strategies. However, we observe that such frameworks struggle with long-term adaptation and fail to generalize to unseen patterns.
  To this end, we introduce \proposed, an LLM-based online time series forecasting framework that leverages a pre-trained LLM as the backbone online forecaster, motivated by its strong few-shot capabilities.
  For efficient online adaptation, \proposed~keeps the LLM frozen and employs compositional spectral prompts grounded in frequency-domain bases to guide the model with the overall distribution of the input, thereby substantially reducing the number of parameters updated during the online phase.
  Specifically, \proposed~decomposes time series into frequency bases and composes the corresponding spectral basis prompts according to their amplitudes, allowing unseen patterns to be represented as new combinations of learned basis prompts.
  Our extensive experiments on real-world datasets demonstrate the superiority and practicality of \proposed~across challenging online scenarios, including extended online phases and cross-dataset settings with substantial distribution shifts.
  Our code is available at ~\url{https://github.com/seungyoon-Choi/CoSPOT}.
\end{abstract}




\renewcommand{\shortauthors}{Seungyoon Choi, Hyunchul Kim, Jae-Gil Lee, \& Chanyoung Park}
\maketitle

\section{Introduction}
Early research in deep learning-based time series forecasting \citep{nie2022time,zhang2023crossformer,zhou2022fedformer,zhou2021informer,wu2021autoformer,wu2022timesnet} primarily focused on batch learning methods utilizing static training and evaluation datasets. However, given the sequential and evolving nature of time series data, shifts in underlying patterns over time are inevitable. Consequently, traditional batch learning approaches often fail to adapt to such changes, while frequent model retraining to accommodate new patterns is both labor-intensive and impractical for real-world applications. To address these challenges, online learning paradigms, which enable models to incrementally update as new data arrive in dynamic environments, have been increasingly explored. 

\begin{figure}[t]
  \centering
  \includegraphics[width=0.9\linewidth]{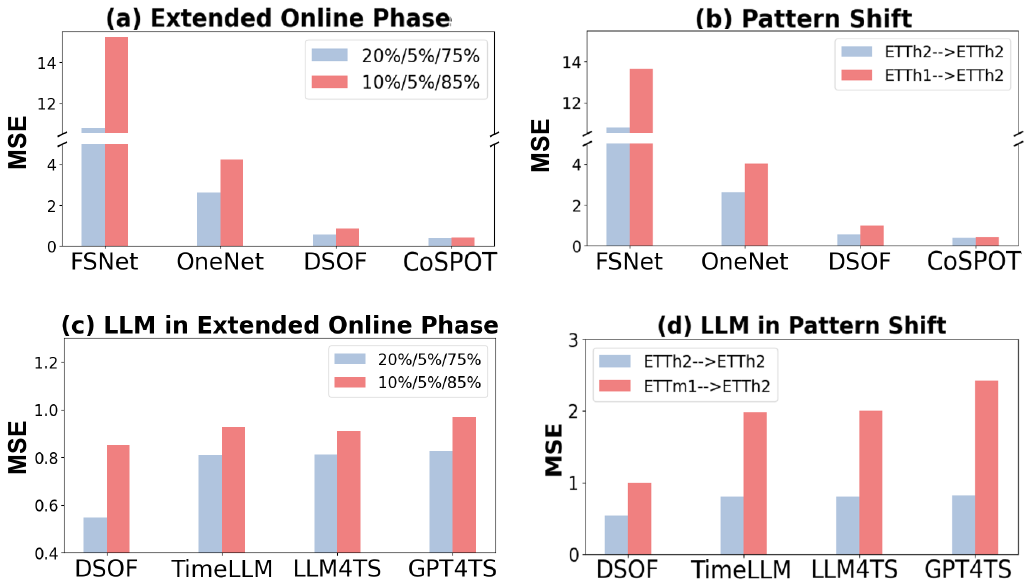}
  \caption{(a) Performance of prior methods and our proposed method (i.e., \proposed) under an extended online phase. (b) Performance of prior methods and \proposed~in a cross-dataset scenario with distribution shifts in pattern. (c \& d) Performance of DSOF and LLM-based time series forecasting methods under an extended online phase and cross-dataset scenario. Note that the ETTh2 dataset is used for (a) and (c).}
  \label{fig: motivation}
  \vspace{-4ex}
\end{figure}

Existing studies on online time series forecasting (OTSF) \citep{pham2022learning, wen2023onenet, laufast} have focused on effective adaptation to evolving data streams. The pioneering work, FSNet \citep{pham2022learning}, addressed rapid adaptation and pattern reuse by employing lightweight per-layer adapters and an associative memory for pattern retrieval. Building on this, OneNet \citep{wen2023onenet} introduced reinforcement learning to dynamically ensemble models based on their real-time performance. Furthermore, DSOF \citep{laufast} identified the information leakage issue in prior setups and proposed a dual-stream (i.e., teacher-student) residual framework to handle delayed adaptation. These methods share a strategy of employing memory buffer-based retrieval to adapt to non-stationary environments.

Despite the recent advancements, the buffer-based design of existing methods faces two fundamental challenges in practical OTSF scenarios.
\textbf{(1) Inadaptability to extended online phases.} 
As new data is continuously streamed in online scenarios, models require sufficient capacity for long-term adaptation to effectively learn from evolving patterns.
As shown in Figure~\ref{fig: motivation} (a), increasing the proportion of the online phase (i.e., test data) leads to significant performance deterioration in previous approaches\footnote{To create a more challenging and extended online phase within the given dataset, we deviate from the typical 20\%/5\%/75\% split used in prior studies and instead adopt a 10\%/5\%/85\% train/validation/test split.}.
This degradation stems from the inherent capacity constraints of memory buffers: as the online period lengthens, the increasing diversity of recurring patterns makes it difficult to explicitly retain them all in the associative memory, leading to catastrophic forgetting.
\textbf{(2) Inadaptability to unseen patterns.} 
In online forecasting scenarios, previously unseen patterns (i.e., distribution shifts in pattern) may emerge during test time (i.e., online phase). Thus, models must possess adaptability to such shifts for effective adjustments. Figure~\ref{fig: motivation}(b) shows experimental results obtained under a setting where the time series patterns in the training and online phases are intentionally made different. Specifically, we used two datasets from the same domain (i.e., ETTh1 and ETTh2)
and conducted experiments under two scenarios: One where model training and online updates are both performed on ETTh2 (in blue), and the other where models are initially trained on ETTh1 dataset, then updated online as ETTh2 data streams in (in red).
We observe that existing OTSF methods degrade substantially on patterns unseen during training, as they rely on associative memory to retrieve patterns similar to the new input. In other words, when an unprecedented pattern appears, retrieval from previously learned patterns becomes unreliable, leading to failure in adaptation under distribution shifts.


To address the challenges of extended online phases and adaptability to unseen patterns, we propose \proposed, a Large Language Model (LLM)-based online time series forecasting framework built upon compositional spectral prompting.
Our design is motivated by the observation that integrating a pre-trained LLM is pivotal for maintaining adaptability over extended online phases.
Recent studies \citep{jin2023time, chang2023llm4ts, zhou2023one} have shown that LLMs can significantly enhance forecasting performance when aligned with time series tasks, particularly in \textit{few-shot} setting. Such transferability of LLMs is especially advantageous in OTSF, as it enables effective adaptation under limited data.
In Figure~\ref{fig: motivation} (c), we compare the state-of-the-art OTSF model that relies solely on a time series model (i.e, DSOF) against those that align the time series backbone with LLMs (i.e., TimeLLM \citep{jin2023time}, LLM4TS \citep{chang2023llm4ts}, and GPT4TS \citep{zhou2023one}) under an extended online phase.
{Surprisingly, we found that although these LLM-based models were not originally designed for online learning\footnote{We adhere to their original training protocols and fine-tune only the output projection in the online phase.}, they demonstrate performance comparable to DSOF as the online phase lengthens.}
However, despite their effectiveness in extended online phases, LLM-based models struggle to generalize effectively to pattern shifts (Figure~\ref{fig: motivation} (d)), limiting their applicability in OTSF. This deficiency underscores the need for continuous adaptation of the model's representational capacity. However, updating the LLM itself is impractical for online scenarios, as the large number of parameters hinders rapid adaptation. To address this, we propose a novel prompting mechanism that alleviates this burden by keeping the LLM frozen, while leveraging a trainable prompt to guide the model’s response to evolving patterns. 
To further enhance the model's generalizability, \proposed~grounds its prompts in frequency-domain representations, which we refer to as spectral prompts.
This distinguishes our approach from prior OTSF methods that rely on time-domain representations, as frequency features are superior at capturing underlying periodic structures \citep{zhou2022fedformer, yi2023frequency}.
Specifically, we decompose the input time series into frequency components using the Discrete Fourier Transform (DFT). 
Each universal frequency basis, which serves as a fundamental building block of time series patterns, is associated with a learnable spectral basis prompt.
These spectral basis prompts are then compositionally aggregated according to the amplitudes of the corresponding frequency components, forming an input-specific compositional spectral prompt fed to the LLM.
This design effectively transforms the adaptation problem: instead of memorizing infinitely many specific patterns, the model represents any unseen pattern as a novel composition of spectral basis prompts that are pre-learned and frozen during the online phase.
Consequently, our framework achieves two key advantages:
(\romannumeral 1) \textit{Generalization}: Newly emerged patterns are represented by recombining spectral basis prompts associated with frequency bases, enabling robust adaptation to distribution shifts.
(\romannumeral 2) \textit{Efficiency}: Since the spectral prompt bank and the LLM are frozen during the online phase, \proposed~achieves inductive adaptation without the computational overhead of large-scale parameter updates.

In this study, we make the following contributions:

\begin{itemize}[leftmargin=10pt]
\item {We identify that existing memory buffer-based OTSF methods struggle with long-term adaptation and fail to generalize to unseen patterns.}
\item {We propose \proposed, the first approach to integrate LLMs into OTSF. \proposed~employs compositional spectral prompting, which constructs input-specific prompts by composing spectral basis prompts grounded in frequency-domain bases.}
\item {Through extensive experiments under various online learning scenarios, we demonstrate that \proposed~consistently outperforms state-of-the-art OTSF methods with extremely few online parameter updates.}
\end{itemize}

\section{Related Works}
\label{related_work}

\noindent\textbf{Online Time Series Forecasting.}
Given the evolving nature of time series data, online forecasting has gained prominence for practical applications \citep{kuznetsov2016time, gultekin2018online, aydore2019dynamic}. Recently, online deep learning models have been proposed to further capture complex patterns within time series data. FSNet \citep{pham2022learning} introduces calibration module to dynamically balance fast adaptation to recent changes with the retention of prior knowledge. OneNet \citep{wen2023onenet} incorporates reinforcement learning to model cross-variable and cross-time concept drifts. Addressing the information leakage issue in previous research, DSOF \citep{laufast} redefines the OTSF setting and proposes a dual-stream mechanism to update model parameters. Nevertheless, prior studies do not explicitly model the patterns in input signals, hindering their ability to adapt to unobserved distributions. Additionally, their explicit storage of pattern information restricts the model's adaptability to extended online phases.

\noindent\textbf{Time Series Forecasting with LLMs.}
Recent advancements in LLMs have prompted researchers to investigate their transferability to forecasting tasks in data-sparse time series domains. LLM4TS \citep{chang2023llm4ts} introduces two-stage fine-tuning approach to leverage LLMs for time series forecasting. GPT4TS \citep{zhou2023one} retrains the positional embeddings and normalization layers of LLMs to preserve pre-trained knowledge while enhancing performance on downstream tasks. Additionally, TimeLLM \citep{jin2023time} employs reprogramming method to align time series data with word embeddings. 
Inspired by the proven adaptability of these models, we leverage LLMs to enable rapid adjustments in online scenarios.

\noindent\textbf{Prompt-based Continual Learning.}
Rehearsal-free continual learning methods leverage the strong general representations of pre-trained models like ViT \citep{dosovitskiy2020image}. By fine-tuning only small, learnable \textit{prompts} for each task, these methods achieve significant memory and computational efficiency, as the core model parameters remain unchanged. VPT \citep{jia2022visual} optimizes a single prompt, L2P \citep{wang2022learning} uses a shared pool of prompts. S-Prompts \citep{wang2022s}, train a unique prompt for each individual task to address catastrophic forgetting.
However, the application of prompt learning to address distribution shifts in the time series domain remains unexplored. Given the continuous nature of time series data, the online learning scenario is more suitable than continual learning, which assumes distinct tasks.
\proposed~is the first to achieve an efficient and scalable prompting crucial for online learning scenarios by encoding knowledge from the underlying frequency bases.

\noindent\textbf{Frequency Analysis in Time Series Forecasting.}
Due to the complex temporal variations in time series data, frequency analysis techniques such as the Discrete Fourier Transform (DFT) and Discrete Wavelet Transform (DWT) are widely used to capture recurring patterns. DFT analyzes global frequency components, while DWT provides localized frequency information at different scales.
FEDformer \citep{zhou2022fedformer} is a representative frequency-domain forecasting model that leverages DFT and DWT through Fourier Enhanced Structure and Wavelet Enhanced Structure, respectively.
However, such frequency-domain approaches mainly exploit frequency coefficients without learning prompt-level knowledge associated with each frequency basis, limiting robustness to unseen patterns under distribution shifts.
In contrast, \proposed~grounds learnable spectral basis prompts in frequency-domain bases and composes them according to the input spectrum, allowing unobserved patterns to be represented as new combinations of learned prompts and improving adaptability in online scenarios.

\section{Preliminaries}
\subsection{Time Series Forecasting}

Let $\mathbf{X}=(x_1, \ldots, x_{N_{data}}) \in \mathbb{R}^{N_{data}\times n}$ be the entire time series with $N_{data}$ observations, where each observation $x_i\in \mathbb{R}^n$ contains $n$ dimensions. The dataset $\mathbf{X}$ is then partitioned into $N_{train}$, $N_{val}$, and, $N_{online}$ time stamps according to predefined ratios, maintaining the chronological order of the data. Given the look-back window of length $L$, denoted as $\mathbf{X}_{i-L+1:i}=(x_{i-L+1}, x_{i-L+2}, \ldots, x_i)$, the objective of time series forecasting is to predict the following $H$ steps (i.e., $\mathbf{X}_{i+1:i+H}$), where the model's prediction at $t=i$ for the next $H$ steps is denoted by $\hat{\mathbf{X}}_{i+1:i+H}=(\hat{x}_{i+1}, \hat{x}_{i+2}, \ldots, \hat{x}_{i+H})=f(\mathbf{X}_{i-L+1:i})$. The objective is to minimize the mean squared error (MSE) between the ground truth and the predicted outputs, i.e., $ \frac{1}{H}\Sigma_{h=1}^{H}||\hat{x}_{i+h}-x_{i+h}||_2^2$.

\subsection{Online Time Series Forecasting}
The OTSF scenario consists of two phases: training phase and online phase. In the training phase, the entire $N_{train}$ time series are utilized to create $(L+H)$ sized time sequences. The objective of the training phase is to let the model to learn the base knowledge through static batch training strategy. The online phase follows the training phase, where $N_{online}$ time steps are streamed sequentially with a moving window of size 1. This mirrors real-world scenarios, and the model is updated in real-time.

\noindent\textbf{Objective and Evaluation Criterion.}
Our ultimate goal is to accurately predict the ground truth by minimizing the cumulative MSE between the ground truth and predicted values over the entire prediction horizon of $H$ steps, using $\textup{MSE}_{online}$ to evaluate performance as follows:
\begin{equation}
\label{eq: online_objective}
\small
\begin{aligned}
\text{MSE}_{\text{online}}
&= \frac{1}{N_{\text{online}} - L - H + 1} \\
&\qquad \sum_{i=N_{\text{train}}+N_{\text{val}}+L}^{N_{\text{data}}-H}
\left\| f(\mathbf{X}_{i-L+1:i})
- \mathbf{X}_{i+1:i+H} \right\|_2^2 .
\end{aligned}
\end{equation}

\section{Proposed Method: \proposed}
\label{sec: method}

In this section, we introduce our proposed method \proposed.
The key components of this framework are as follows: (1) leveraging a pre-trained LLM together with a time series backbone and textual recent information to enhance adaptability in data-scarce online scenarios (Sec.~\ref{sec: text_description}), and (2) compositional spectral prompting, which composes frequency-grounded spectral basis prompts to robustly and efficiently adapt to pattern shifts
(Sec.~\ref{sec: pattern_embedding}). Overall framework of \proposed~is shown in Figure~\ref{fig: architecture}.

\subsection{Enhancing model adaptability using a pre-trained LLM}
\label{sec: text_description}

To enable effective and rapid adaptation in data-scarce online scenarios, \proposed~leverages a pre-trained LLM together with a time-series backbone. The LLM remains frozen throughout the training and online phases, serving as a stable knowledge source, while the time-series backbone encodes the input sequence. This design exploits the LLM's transferability to support adaptation when only limited online observations are available.
However, online forecasting also requires contextual information about recent dynamics, which may not be sufficiently captured from numerical inputs alone. To this end, we introduce a text description as an additional modality, allowing recent contextual cues to be provided to the LLM in natural language.
Given the input time series $\mathbf{X}$, the text description (i.e., $text_{\mathbf{X}}$) contains task details, dataset information, recent values, and recent frequency information. This text description provides short-term contextual cues, complementing the compositional spectral prompt in Section~\ref{sec: pattern_embedding}, which captures the overall distribution of the input. The recent values summarize the latest time-domain variation, while the recent frequency information captures local frequency changes near the current time point. Although Short-Time Fourier Transform (STFT) can provide localized frequency information, its fixed window size limits its ability to capture non-stationary recent dynamics due to the trade-off between time and frequency resolution. Hence, we use the Discrete Wavelet Transform (DWT), whose adaptive time-frequency resolution is better suited for capturing recent variations in non-stationary signals.
Given the input time series $\mathbf{X}$, the equation of DWT using a scaling function $\phi$ and a wavelet function $\psi$ is as follows:
\begin{equation}
\small
\label{eq: DWT}
    \mathbf{A}_j[k] = \sum_{n}\mathbf{X}[n]\phi_{j,k}[n], \quad \mathbf{D}_j[k]= \sum_{n}\mathbf{X}[n]\psi_{j,k}[n],
\vspace{-2.5ex}
\end{equation}
where $\mathbf{X}[n]$ refers to the $n$-th index in the time series $\mathbf{X}$, $\mathbf{A}_j[k] \,\text{and}\,\mathbf{D}_j[k]$ refer to the approximation and detail coefficient at level $j$, respectively, and $\phi_{j,k}[n] \,\text{and}\, \psi_{j,k}[n]$ are the scaling and wavelet functions at level $j$, respectively. After the decomposition, the time series is passed through a filter bank that separates the low-pass and high-pass components, and downsampling is performed as follows:
\begin{equation}
\small
\label{eq: appro_detail}
    \mathbf{A}_{j+1}[k] = \sum_{n}h[n-2k]\mathbf{A}_j[n], \quad \mathbf{D}_{j+1}[k]= \sum_{n}g[n-2k]\mathbf{A}_j[n],
\end{equation}
where $h[n] \,\text{and}\, g[n]$ refer the low- and high-pass filters, respectively. Through this process, we utilize $\mathbf{A}_j[-1]$ to provide the model with the most recent frequency information where $j$ is a hyperparameter. 

\begin{figure}[!t]  
  \centering        
  
  \includegraphics[width=1.0\linewidth]{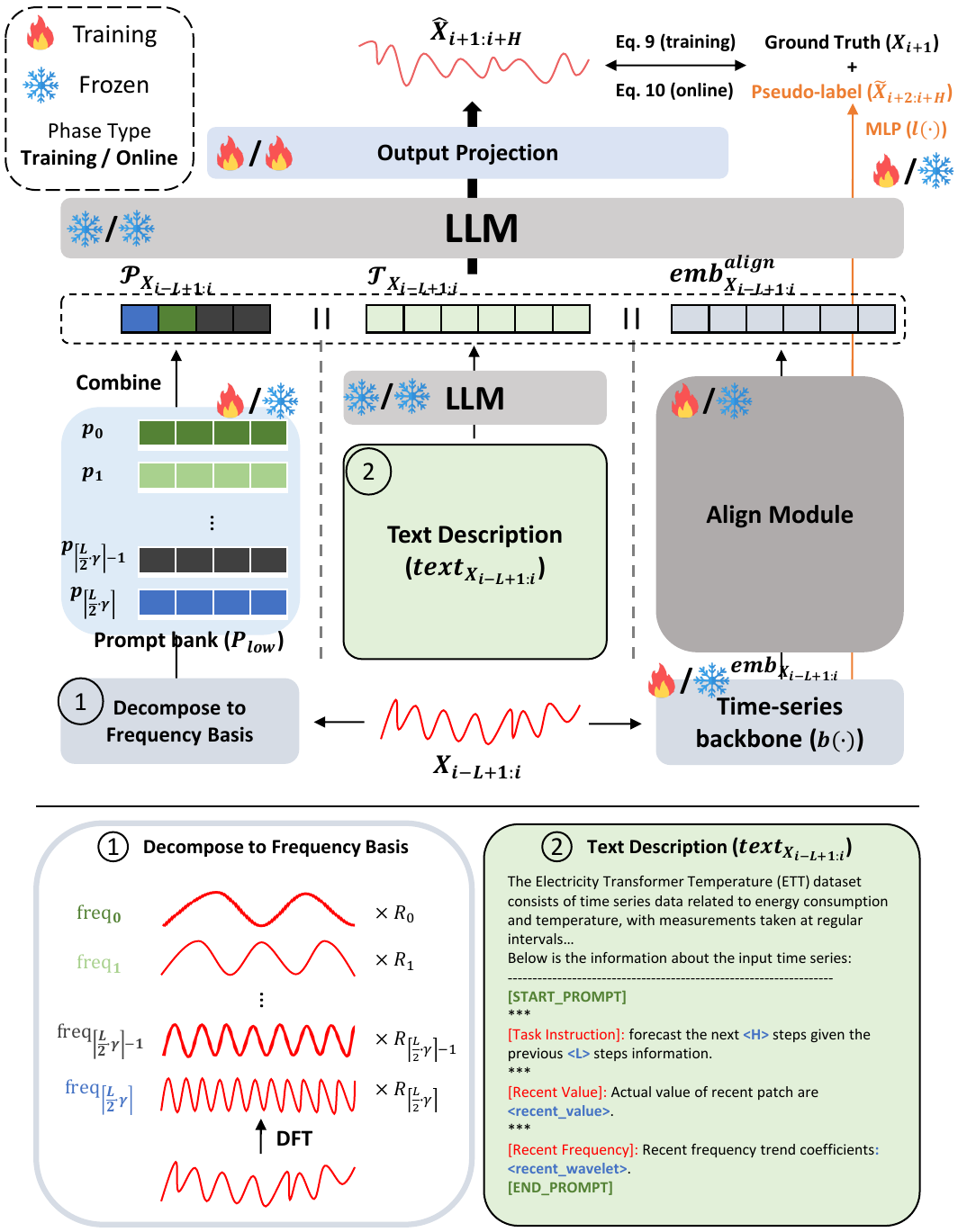}

  \caption{Overall model framework. Given the input time series $\mathbf{X}_{i-L+1:i}$, the aligned embedding (i.e., $emb_{\mathbf{X}_{i-L+1:i}}$), embedded text description (i.e., $\mathcal{T}_{\mathbf{X}_{i-L+1:i}}$), and compositional spectral prompt (i.e., $\mathcal{P}_{\mathbf{X}_{i-L+1:i}}$) are provided as input to the LLM. The representation computed by the LLM is passed through the output projection layer to produce the final prediction, i.e., $\mathbf{\hat{X}}_{i+1:i+H}$.}
  \label{fig: architecture}
  \vspace{-4.5ex}
  
\end{figure}

An example of the text description is shown in Figure~\ref{fig: architecture}. The text description is first processed through the pre-trained LLM's tokenizer. The resulting token IDs are then passed through the frozen LLM's input embedding layer to retrieve their corresponding token embeddings. This sequence of text token embeddings is the resulting embedding (denoted as $\mathcal{T}_{\mathbf{X}}$) that is used as input for the final prediction.
Providing text descriptions enriches data in data-scarce online scenarios, offering recent information that aids effective adaptation, all without requiring additional training.

\subsection{Robust Adaptation via Compositional Spectral Prompting}
\label{sec: pattern_embedding}

While the frozen LLM integrated with text descriptions provides a strong foundation for adaptability, it still faces challenges when encountering completely unseen patterns that deviate significantly from the patterns learned during the training phase. However, updating the LLM itself is impractical for online scenarios, as the large number of parameters hinders rapid adaptation. Therefore, we need an efficient mechanism to provide the frozen model with inductive bias about these evolving structures.
To address this, we introduce \textit{compositional spectral prompting}, which provides the frozen model with reliable guidance on the overall data distribution. Here, \textit{spectral} indicates that the prompts are grounded in the frequency-domain structure of time series. 
We leverage the frequency domain because, compared with the time domain, it more effectively isolates underlying pattern components that are often entangled in raw temporal signals, making it well suited for representing evolving time-series patterns\footnote{Section~\ref{sec: time_prompting} elaborates on the comparative advantages of the frequency-domain approach, highlighting its enhanced suitability for robust pattern representation compared to the time-domain approach.}. Instead of directly storing complex patterns, which renders prior memory-based methods~\citep{pham2022learning, wen2023onenet, laufast} unreliable when facing unseen patterns, \proposed~decomposes time series into frequency bases and represents each input as a composition of learnable spectral basis prompts. This design allows newly emerging patterns to be expressed through new combinations of pre-learned prompts, without updating the LLM during the online phase. Specifically, we use the DFT to decompose the input time series $\mathbf{X}$ of length $L$. The DFT converts the sequence from the time domain to the frequency domain, while its inverse (IDFT) converts it back. Their expressions are as follows:

\begin{equation}
\small
\label{eq: DFT}
    \mathcal{F}(k) = DFT(\mathbf{X}) = \sum_{n=0}^{L-1}\mathbf{X}[n]\textup{exp}\Big(-i\frac{2\pi kn}{L}\Big), \quad k=0,1,\ldots,L-1,
\end{equation}
\begin{equation}
\small
\label{eq: IDFT}
    \mathbf{X}[n] = IDFT(\mathcal{F})=\frac{1}{L}\sum_{k=0}^{L-1}\mathcal{F}(k)\textup{exp}\Big(i\frac{2\pi kn}{L}\Big), \quad n=0,1,\ldots, L-1,
\end{equation}
where $\mathcal{F}$ refers to the frequency spectrum of the input and $i$ represents the imaginary unit. From the perspective of frequency basis, assuming that $L$ is even, both the DFT and IDFT can be represented using $\frac{L}{2}+1$ orthogonal cosine-based frequency basis because the DFT of a real-valued signal exhibits Hermitian symmetry. 
Thus, the IDFT can be rewritten as follows:

\begin{equation}
\label{eq: IDFT_v2}
\small
\begin{aligned}
\mathbf{X}[n]
&= \frac{1}{L}\sum_{k=0}^{\frac{L}{2}}
\Big( \mathbf{R}_k \cdot 
\cos\Big( \frac{2\pi kn}{L} - \phi \Big) \Big) \\
&\qquad = \frac{1}{L}\sum_{k=0}^{\frac{L}{2}}
\Big( \mathbf{R}_k \cdot \text{freq}_k \Big),
\quad n = 0, 1, \ldots, L-1 .
\end{aligned}
\end{equation}

\noindent where $\text{freq}_k$ and $\mathbf{R}_k\in\mathbb{R}$ denote the basis of the $k$-th frequency and its amplitude, respectively. 

Building on this formulation, which reconstructs the time series as a weighted combination of frequency bases, we introduce a spectral prompt bank that associates each frequency basis with a learnable spectral basis prompt. Specifically, each spectral basis prompt encodes the characteristic temporal pattern corresponding to its frequency basis.
Let $\mathbf{P}=[\mathbf{p}_0, \ldots, \mathbf{p}_{\frac{L}{2}}]\in\mathbb{R}^{(\frac{L}{2}+1) \times d}$ denote the spectral prompt bank, where $\mathbf{p}_k$ represents the learnable spectral basis prompt corresponding to the $k$-th frequency basis.
However, learning knowledge for all frequency bases is not effective in capturing the overall pattern of the given time series. That is, high frequencies represent rapidly oscillating periodicities compared to low frequencies, and therefore, they do not capture the overall pattern information. Hence, in time series analysis, high frequencies are often treated as noise, which is why low-pass filtering techniques \citep{zhou2022film,xu2023fits} are widely used. Accordingly, to effectively capture the overall pattern while removing noise, we introduce a hyperparameter $\gamma\in[0,1]$ to eliminate the high-frequency bases, i.e., $\mathbf{P}_{low}=[\mathbf{p}_0, \ldots, \mathbf{p}_{\left \lceil \frac{L}{2}\cdot \gamma \right \rceil}]\in\mathbb{R}^{\left \lceil (\frac{L}{2}\cdot \gamma+1)\right \rceil \times d}$. 
We then construct an input-specific compositional spectral prompt by weighting each spectral basis prompt according to the amplitude of its corresponding frequency component, which provides the LLM with guidance on the overall distribution (i.e., overall pattern) of the input:

\begin{equation}
\label{eq: pattern_embedding}
    \mathcal{P}_{\mathbf{X}} = \textup{Concat}\Big(\mathbf{R}_0 \cdot \mathbf{p}_0, \mathbf{R}_1 \cdot \mathbf{p}_1, \ldots, \mathbf{R}_{\left \lceil \frac{L}{2}\cdot \gamma \right \rceil} \cdot \mathbf{p}_{\left \lceil \frac{L}{2}\cdot \gamma \right \rceil}\Big) \in \mathbb{R}^{\left \lceil (\frac{L}{2}\cdot \gamma+1)\right \rceil \times d},
\vspace{-0.5ex}
\end{equation}
where $\mathcal{P}_{\mathbf{X}}$ refers the compositional spectral prompt of the input time series $\mathbf{X}$ to be provided to the model. 
Through compositional spectral prompting, the model explicitly captures pattern-level information under continuous distribution shifts.
Even when previously unseen patterns emerge in the online phase, the model can effectively represent and adapt to them 
by compositionally recombining the knowledge encoded in spectral basis prompts
during the training phase, without requiring any additional training.\footnote{In Section~\ref{sec: generalizability}, we observe that the frequency basis-driven prompt bank, trained only during the training phase, can effectively adapt to unseen patterns that emerge in the online phase.}
Moreover,
since newly emerging patterns can be expressed as compositions of a finite set of spectral basis prompts, the model maintains its memory efficiency without degradation.

\vspace{-1ex}
\subsection{Overall Framework}
\label{sec: overall_framework}

Figure~\ref{fig: architecture} shows the overall framework, and the pseudo code can be found in Algorithm \ref{alg:algorithm1}.


  
  
  
  

\smallskip
\noindent\textbf{Align Module.}
In OTSF scenarios, effectively aligning continuous time series data with discrete token-based LLMs is crucial yet challenging. Since pre-trained LLMs lack inherent knowledge of time series patterns, prior studies \citep{jin2023time,chang2023llm4ts,zhou2023one} focus on aligning two modalities (i.e., time series and language) to leverage the knowledge within LLMs, enabling accurate, data-efficient, and task-agnostic forecasting.
As the goal of this study is to enhance the LLM's ability in the online scenario, not to focus on the align module itself, we utilize a pre-developed align module\citep{jin2023time,chang2023llm4ts}.
The align module aligns the representation of the time series computed by the time series backbone (i.e., $emb_{\mathbf{X}} = b(\mathbf{X})$, where $b(\cdot)$ is the time series backbone) with the natural language modality, and outputs the resulting aligned embeddings (i.e., $emb^{align}_{\mathbf{X}}$).

\smallskip
\noindent\textbf{Utilizing a pre-trained LLM.}
The input to the LLM is formed by concatenating the compositional spectral prompt (i.e., $\mathcal{P}$), embedded text description (i.e., $\mathcal{T}$), and aligned time series embedding (i.e., $emb^{align}$).
Within the text description, we utilize special tokens (e.g., \texttt{[START PROMPT]} and \texttt{[END PROMPT]}) as delimiters, enabling the LLM to clearly discern the boundaries between these heterogenous modalities.
The concatenated embedding sequence is input to the LLM's transformer layers, all of which remain frozen.
The last hidden state representation of the LLM serves as the time series representation, which is flattened and linearly projected to generate the final forecast.


\smallskip
\noindent\textbf{Training Phase.}
The training phase serves to learn the overall base knowledge before entering the online phase. Therefore, during the training phase, except for the pre-trained LLM, we train the time series backbone network, align module, spectral prompt bank, and output projection layer by minimizing the following objective:
\vspace{-1ex}
\begin{equation}
\label{eq: training_loss}
    \mathcal{L}_{training} =\frac{1}{N_{train}-L-H+1}\sum_{i=L}^{N_{train}-H}||f(\mathbf{X}_{i-L+1:i})-\mathbf{X}_{i+1:i+H}||_2^2.
\vspace{-1ex}
\end{equation}
where $f(\cdot)$ is the overall framework of~\proposed.


\noindent\textbf{Online Phase.}
In the online phase, only the output projection layer is tuned to match the streaming data distribution, while all other parameters, including the pre-trained LLM, time series backbone network, align module, and spectral prompt bank, remain frozen.\footnote{We emphasize that \proposed~is an efficient framework despite utilizing an LLM, as the number of parameters updated during the online phase is extremely small. Please see Section~\ref{sec: online_update} for a detailed analysis.}
According to \citep{laufast}, when the prediction horizon $H$ is greater than 1, calculating the loss using the ground truth for all $H$ time steps ($\mathbf{X}_{i+1:i+H}$) at each moving window step for model updates leads to information leakage. To avoid this, the moving window should be extended to $H$ steps rather than updating at each step, which however introduces an update delay and hinders effective adaptation. Hence, we employ a pseudo-labeling technique while keeping updates at each step.
When calculating the loss for the model's output at $t=i$ to update the model, only the ground truth for the immediate next time point, i.e., $\mathbf{X}_{i+1}$ is used, and pseudo-labels are generated and utilized for the remaining time steps (i.e.,$\mathbf{\tilde{X}}_{i+2:i+H}$). We propagate the representation of the time series backbone network (i.e., $b(\cdot)$) through a linear layer (i.e., $l(\cdot)$) to project it into the output space and obtain the pseudo-label. This linear layer is utilized while being frozen during the online phase, and is trained along with Equation~\ref{eq: training_loss} during the training phase using MSE loss. Therefore, Equation~\ref{eq: training_loss} is modified as follows:
\vspace{-1ex}
\begin{equation}
\label{eq: training_loss_v2}
\begin{aligned}
\mathcal{L}^{+}_{\text{training}}
= \frac{1}{N_{\text{train}} - L - H + 1}
&\sum_{i=L}^{N_{\text{train}}-H}
\Big( \left\| f(\mathbf{X}_{i-L+1:i})
- \mathbf{X}_{i+1:i+H} \right\|_2^2 \\
&\qquad + \left\| l\!\left( b(\mathbf{X}_{i-L+1:i}) \right)
- \mathbf{X}_{i+1:i+H} \right\|_2^2 \Big).
\end{aligned}
\end{equation}

\noindent where $l(\cdot)$ denotes the pseudo-label projection layer which projects the time series representation into the output space. 
During the online phase, we use the frozen $b(\cdot)$ and $l(\cdot)$, which are trained in the training phase, to generate pseudo-labels:  $\mathbf{\tilde{X}}_{i+2:i+H}=l(b(\mathbf{X}_{i-L+1:i}))[1:]$. These pseudo-labels are then employed to tune the model.
To mitigate the impact of prediction errors and pseudo-labels as the forecast horizon extends from the current observation,
we apply a geometric decay factor $\delta\in[0,1]$ to the online loss as follows:
\vspace{-2ex}

\begin{equation}
\label{eq: online_loss}
\begin{aligned}
\mathcal{L}_{\text{online}}
&\hspace{-0.1em}= \frac{1}{N_{\text{online}} - L - H + 1}
\sum_{i=N_{\text{train}}+N_{\text{val}}+L}^{N_{\text{data}}-H} \\[0.2ex]
&\qquad\qquad
\Big( \frac{1}{H}
\sum_{h=1}^{H} \delta^{h-1}
\left\| \hat{\mathbf{X}}_{i+h}
- \bar{\mathbf{X}}_{i+h} \right\|_2^2 \Big).
\end{aligned}
\vspace{-1ex}
\end{equation}

\vspace{-0.1ex}
\noindent where $\mathbf{\overline{X}}_{i+1:i+H}=\textup{Concat}(\mathbf{X}_{i+1}, \mathbf{\tilde{X}}_{i+2:i+H})$ is a concatenated sequence of the ground truth and pseudo-labels. Finally, the model's online performance is evaluated using Equation~\ref{eq: online_objective}.

\noindent\textbf{Pseudocode.}
\label{sec: pseudo_code}
Algorithm~\ref{alg:algorithm1} describes the training and online phases of \proposed.

\begin{algorithm}[h]
\caption{Pseudocode for training and online phases of \proposed}
\label{alg:algorithm1}
\small
\begin{algorithmic}[1]
\State \textbf{Input:} Time Series $\mathbf{X}\in\mathbb{R}^{N_{data}}$, Pre-trained LLM, Batch Size, Align Module
\State \textbf{Output:} Time Series Backbone Network $b(\cdot)$, Spectral Prompt Bank $\mathbf{P}$, Output Projection Layer

\State{}
\State \textbf{\# Training Phase}

\State Freeze the Pre-trained LLM

\For{i in range($\tfrac{N_{train}}{\textup{Batch Size}}$)}
    \State $\mathbf{X}^{(i)}\in\mathbb{R}^{\textup{Batch Size}\times L}$
    \Comment{$i$-th batch for batch training}
    \State{}
    \State \textbf{\# Aligning}
    \State $emb_{\mathbf{X}^{(i)}}=b(\mathbf{X}^{(i)})$
    \State Generate $emb_{\mathbf{X}^{(i)}}^{align}$ using the Align Module
    \State{}
    \State \textbf{\# Text Description}
    \State Generate $\mathcal{T}_{\mathbf{X}^{(i)}}$ using $text_{\mathbf{X}^{(i)}}$
    \State{}
    \State \textbf{\# Compositional Spectral Prompt}
    \State Decompose $\mathbf{X}^{(i)}$ via DFT into bases and amplitudes
    \Comment{Eq. \ref{eq: IDFT_v2}}
    \State Compose $\mathcal{P}_{\mathbf{X}^{(i)}}$ from spectral basis prompts
    \Comment{Eq. \ref{eq: pattern_embedding}}

    \State{}
    \State Feed $[\mathcal{P}_{\mathbf{X}^{(i)}};\mathcal{T}_{\mathbf{X}^{(i)}};emb_{\mathbf{X}^{(i)}}^{align}]$ into the LLM
    \State The Pre-trained LLM's representation is projected to obtain $\hat{\mathbf{X}}^{(i)}$

    \State{}
    \State Train all parameters except the Pre-trained LLM 
    \Comment{Eq. \ref{eq: training_loss_v2}}
\EndFor

\State{}
\State \textbf{\# Online Phase}

\State Freeze all parameters except those in the Output Projection Layer
\For{i in range($N_{online}$)}
    \State $\mathbf{X}^{(i)}\in\mathbb{R}^{L}$
    \Comment{$i$-th data instance}
    \State{}
    \State \textbf{\# Aligning}
    \State $emb_{\mathbf{X}^{(i)}}=b(\mathbf{X}^{(i)})$
    \State Generate $emb_{\mathbf{X}^{(i)}}^{align}$ using the Align Module
    \State{}
    \State \textbf{\# Text Description}
    \State Generate $\mathcal{T}_{\mathbf{X}^{(i)}}$ using $text_{\mathbf{X}^{(i)}}$
    \State{}
    \State \textbf{\# Compositional Spectral Prompt}
    \State Decompose $\mathbf{X}^{(i)}$ via DFT into bases and amplitudes
    \Comment{Eq. \ref{eq: IDFT_v2}}
    \State Compose $\mathcal{P}_{\mathbf{X}^{(i)}}$ from spectral basis prompts
    \Comment{Eq. \ref{eq: pattern_embedding}}

    \State{}
    \State Feed $[\mathcal{P}_{\mathbf{X}^{(i)}};\mathcal{T}_{\mathbf{X}^{(i)}};emb_{\mathbf{X}^{(i)}}^{align}]$ into the LLM
    \State The Pre-trained LLM's representation is projected to obtain $\hat{\mathbf{X}}^{(i)}$
    \State{}
    \State Update the Output Projection Layer using $\mathcal{L}_{online}$
    \Comment{Eq. \ref{eq: online_loss}}
\EndFor
\State{}
\State Evaluate the online phase
\Comment{Eq. \ref{eq: online_objective}}

\end{algorithmic}
\end{algorithm}

\vspace{-1.25ex}
\section{Experiments}
\label{sec: experiments}

\noindent\textbf{Datasets.}
Following prior studies \citep{pham2022learning, wen2023onenet, laufast}, we evaluate our method on five widely used time series forecasting benchmarks from various domains, splitting each into training, validation, and testing sets with a 20\%, 5\%, and 75\% ratio, respectively.
\textbf{ETT}\footnote{https://github.com/zhouhaoyi/ETDataset} contains electricity load and oil temperature data collected at 15-minute (ETTm1, ETTm2) and hourly (ETTh1, ETTh2) intervals, each with 6 covariates.
\textbf{Weather}\footnote{https://www.ncei.noaa.gov/data/local-climatological-data/} consists of 21 meteorological variables recorded at hourly intervals, covering diverse climate features such as temperature, humidity, and wind speed.
\textbf{ECL}\footnote{https://archive.ics.uci.edu/ml/datasets/ElectricityLoadDiagrams20112014} comprises hourly electricity consumption records from 321 clients over two years.
\textbf{Traffic}\footnote{https://pems.dot.ca.gov/} contains occupancy rate measurements from 862 freeway sensors in the San Francisco Bay Area, recorded at 5-minute intervals.
\textbf{Exchange Rate}\footnote{https://github.com/laiguokun/multivariate-time-series-data} contains daily exchange-rate records across multiple countries. Compared with the other benchmarks, Exchange Rate exhibits less pronounced periodicity and is more influenced by random economic and market events, providing a challenging testbed for evaluating robustness beyond strongly seasonal patterns.

\smallskip
\noindent\textbf{Baselines.}
We utilize various deep learning-based time series forecasting models as baselines, categorized into four groups.
\textit{Static forecasting models}: \textbf{DLinear} \citep{zeng2023transformers} decomposes time series into trend and seasonal components and applies linear layers to each component. \textbf{PatchTST} \citep{nie2022time} applies patch-based Transformer encoding to preserve local temporal patterns. \textbf{iTransformer} \citep{liu2023itransformer} attends along the feature dimension to capture cross-variable interactions. \textbf{TimeMixer} \citep{wang2024timemixer} introduces a multiscale-mixing architecture that decomposes and mixes seasonal and trend components across different sampling scales.
\textit{Time series foundation models}: \textbf{Chronos-2} \citep{ansari2025chronos} is a time series foundation model that supports zero-shot univariate, multivariate, and covariate-informed forecasting through a unified architecture. \textbf{TimesFM} \citep{das2023decoder} is a decoder-only foundation model pre-trained on large-scale time series corpora for zero-shot forecasting. 
\textit{LLM-based models}: \textbf{LLM4TS} \citep{chang2023llm4ts} aligns time series representations with pre-trained LLMs through a two-stage fine-tuning strategy for forecasting. \textbf{GPT4TS} \citep{zhou2023one} repurposes frozen GPT-style Transformers by tuning only lightweight components for time series prediction. \textbf{Time-LLM} \citep{jin2023time} reprograms time series into text prototypes with a Prompt-as-Prefix strategy.
\textit{Online time series forecasting (OTSF) models}: \textbf{FSNet} \citep{pham2022learning} captures short- and long-term patterns via dual fast and slow learners. \textbf{OneNet} \citep{wen2023onenet} adaptively weights an online ensemble based on recent performance. \textbf{DSOF} \citep{laufast} introduces a dual-stream framework that updates parameters through distinct short- and long-term temporal contexts.

\noindent\textbf{Implementation Details.}
Consistent with previous studies \citep{pham2022learning, wen2023onenet, laufast}, we set prediction length $H$ to 1, 24, and 48, with a lookback length $L$ of 96. We utilize PatchTST \citep{nie2022time} as the time series backbone network $b(\cdot)$ and Llama-7B \citep{touvron2023llama} as the default LLM unless stated otherwise. The align module of Time-LLM \citep{jin2023time} is used as the default align module. The evaluation metrics include mean square error (MSE) and mean absolute error (MAE). Our method is implemented with Python 3.11 and PyTorch 2.2.2. We use the AdamW optimizer, training for 10 epochs during the training phase and performing one-step updates per data instance during the online phase. Key hyperparameters are set as follows: $\gamma = 0.3$ (Eq.~\ref{eq: pattern_embedding}) for low-pass filtering, wavelet decomposition level $j=2$ (Eq.~\ref{eq: appro_detail}), and $\delta=0.8$ (Eq.~\ref{eq: online_loss}) to assign stronger supervision to near-future values. All experiments are conducted on a 48GB NVIDIA RTX A6000.

\vspace{-1ex}

\begin{table*}[t]
\vspace{-1ex}
\caption{{Comparison of MSE and MAE results in OTSF for predicting 1, 24, and 48 prediction horizon (i.e., $H$) with a lookback length $L=96$ (Best: bold red, the second-best: underlined in blue).}}
\vspace{-2ex}
\centering
\renewcommand{\arraystretch}{1}
    \resizebox{0.99\linewidth}{!}{
\begin{tabular}{cccccccccccccccccccccccccccc}
\toprule
 & & \multicolumn{2}{c}{DLinear} & \multicolumn{2}{c}{PatchTST} & \multicolumn{2}{c}{iTransformer} & \multicolumn{2}{c}{TimeMixer} & \multicolumn{2}{c}{Chronos-2} & \multicolumn{2}{c}{TimesFM} & \multicolumn{2}{c}{LLM4TS} & \multicolumn{2}{c}{GPT4TS} & \multicolumn{2}{c}{Time-LLM} & \multicolumn{2}{c}{FSNet} & \multicolumn{2}{c}{OneNet} & \multicolumn{2}{c}{DSOF} & \multicolumn{2}{c}{\proposed} \\
\cline{3-28}
&  $H$ & MSE & MAE & MSE & MAE & MSE & MAE & MSE & MAE & MSE & MAE & MSE & MAE & MSE & MAE & MSE & MAE & MSE & MAE & MSE & MAE & MSE & MAE & MSE & MAE & MSE & MAE \\
\hline
\multirow{3}{*}{\rotatebox{90}{ETTh1}} & 1 & \textcolor{blue}{\underline{0.502}} & \textcolor{blue}{\underline{0.609}} & 0.779 & 0.828 & 0.993 & 0.976 & 0.557 & 0.706 & 0.579 & 0.681 & 0.621 & 0.708 & 1.436 & 1.058 & 1.539 & 1.141 & 1.382 & 1.076 & 13.26 & 3.441 & 4.023 & 1.956 & 0.802 & 0.857 & \textbf{\textcolor{red}{0.425}} & \textbf{\textcolor{red}{0.601}} \\
                        & 24 & 2.333 & 1.427 & 3.797 & 1.849 & 3.028 & 1.690 & 2.719 & 1.528 & 2.484 & 1.396 & 2.418 & 1.455 & 3.155 & 1.623 & 3.028 & 1.640 & 2.947 & 1.417 & 19.33 & 4.097 & 9.001 & 2.801 & \textcolor{blue}{\underline{2.311}} & \textcolor{blue}{\underline{1.372}} & \textbf{\textcolor{red}{1.217}} & \textbf{\textcolor{red}{0.993}} \\
                        & 48 & 2.802 & 1.553 & 5.135 & 2.066 & 3.998 & 1.929 & 3.274 & 1.699 & \textcolor{blue}{\underline{2.748}} & \textcolor{blue}{\underline{1.458}} & 2.994 & 1.630 & 6.256 & 2.312 & 6.738 & 2.397 & 5.997 & 2.049 & 25.82 & 4.881 & 12.21 & 3.334 & 4.233 & 1.973 & \textbf{\textcolor{red}{1.482}} & \textbf{\textcolor{red}{1.073}} \\
\hline
\multirow{3}{*}{\rotatebox{90}{ETTh2}} & 1 & 0.468 & 0.643 & 0.895 & 0.924 & 0.874 & 0.888 & \textcolor{blue}{\underline{0.439}} & 0.563 & 0.476 & 0.590 & 0.597 & 0.572 & 0.811 & 0.875 & 0.827 & 0.829 & 0.810 & 0.836 & 10.78 &  3.083 & 2.634 & 1.531 & 0.549 & \textcolor{blue}{\underline{0.537}} & \textbf{\textcolor{red}{0.398}} & \textbf{\textcolor{red}{0.423}} \\
                        & 24 & 2.187 & 1.338 & 4.885 & 2.002 & 2.688 & 1.451 & 1.938 & 1.296 & \textcolor{blue}{\underline{1.343}} & \textcolor{blue}{\underline{1.056}} & 1.419 & 1.091 & 3.023 & 1.538 & 2.887 & 1.509 & 2.381 & 1.503 & 17.99 & 4.041 & 7.019 & 2.338 & 2.179 & 1.410 & \textbf{\textcolor{red}{0.932}} & \textbf{\textcolor{red}{0.607}} \\
                        & 48 & 2.349 & 1.412 & 6.368 & 2.349 & 3.768 & 1.541 & 2.216 & 1.398 & \textcolor{blue}{\underline{2.103}} & \textcolor{blue}{\underline{1.250}} & 2.108 & 1.252 & 5.411 & 2.026 & 5.510 & 2.237 & 5.231 & 2.087 & 22.79 & 4.673 & 9.930 & 3.019 & 4.003 & 1.991 & \textbf{\textcolor{red}{1.245}} & \textbf{\textcolor{red}{0.699}} \\
\hline
\multirow{3}{*}{\rotatebox{90}{ETTm1}} & 1 & 0.132 & 0.312 & 0.135 & 0.352 & 0.183 & 0.397 & 0.137 & 0.341 & 0.145 & 0.224 & 0.173 & 0.260 & 0.371 & 0.489 & 0.387 & 0.522 & 0.309 & 0.506 & 0.190 & 0.231 & 0.154 & 0.271 & \textcolor{blue}{\underline{0.096}} & \textcolor{blue}{\underline{0.142}} & \textbf{\textcolor{red}{0.083}} & \textbf{\textcolor{red}{0.128}} \\
                        & 24 & 0.618 & 0.771 & 1.102 & 1.029 & 1.101 & 1.009 & 0.602 & 0.726 & 0.717 & 0.600 & 0.912 & 0.731 & 0.702 & 0.707 & 0.711 & 0.741 & 0.671 & 0.795 & 1.520 & 1.202 & 1.103 & 1.003 & \textcolor{blue}{\underline{0.412}} & \textcolor{blue}{\underline{0.452}} & \textbf{\textcolor{red}{0.372}} & \textbf{\textcolor{red}{0.403}} \\
                        & 48 & 0.829 & 0.894 & 2.083 & 1.243 & 1.298 & 1.039 & 0.811 & 0.882 & 1.052 & 0.781 & 1.152 & 0.890 & 0.935 & 0.846 & 0.933 & 0.886 & 0.897 & 0.836 & 2.283 & 1.390 & 1.492 & 1.21 & \textcolor{blue}{\underline{0.559}} & \textcolor{blue}{\underline{0.540}} & \textbf{\textcolor{red}{0.451}} & \textbf{\textcolor{red}{0.515}} \\
                        \hline
\multirow{3}{*}{\rotatebox{90}{ETTm2}} & 1 & 0.111 & 0.303 & 0.113 & 0.296 & 0.137 & 0.350 & 0.114 & 0.326 & 0.107 & 0.257 & 0.126 & 0.305 & 0.302 & 0.525 & 0.321 & 0.536 & 0.187 & 0.333 & 0.127 & 0.326 & 0.111 & 0.297 & \textcolor{blue}{\underline{0.066}} & \textcolor{blue}{\underline{0.219}} & \textbf{\textcolor{red}{0.052}} & \textbf{\textcolor{red}{0.208}} \\
                        & 24 & 0.589 & 0.667 & 1.035 & 0.917 & 0.799 & 0.853 & 0.525 & 0.688 & 0.647 & 0.774 & 0.624 & 0.710 & 0.631 & 0.714 & 0.587 & 0.706 & 0.499 & 0.624 & 1.180 & 0.976 & 0.603 & 0.716 & \textcolor{blue}{\underline{0.348}} & \textcolor{blue}{\underline{0.519}} & \textbf{\textcolor{red}{0.287}} & \textbf{\textcolor{red}{0.517}} \\
                        & 48 & 0.809 & 0.794 & 1.802 & 1.202 & 1.003 & 0.950 & 0.784 & 0.854 & 0.815 & 0.823 & 0.846 & 0.880 & 0.775 & 0.833 & 0.713 & 0.749 & 0.589 & 0.707 & 1.847 & 1.194 & 0.843 & 0.881 & \textcolor{blue}{\underline{0.412}} & \textcolor{blue}{\underline{0.628}} & \textbf{\textcolor{red}{0.348}} & \textbf{\textcolor{red}{0.539}} \\
\hline
\multirow{3}{*}{\rotatebox{90}{WTH}}   & 1 & 0.359 & 0.499 & 0.521 & 0.691 & 0.491 & 0.607 & 0.338 & 0.543 & \textcolor{blue}{\underline{0.205}} & \textcolor{blue}{\underline{0.352}} & 0.360 & 0.400 & 0.489 & 0.659 & 0.493 & 0.602 & 0.488 & 0.649 & 0.731 & 0.754 & 0.501 & 0.536 & 0.298 & 0.398 & \textbf{\textcolor{red}{0.052}} & \textbf{\textcolor{red}{0.114}} \\
                        & 24 & 1.219 & 1.008 & 1.504 & 1.208 & 1.482 & 1.017 & 1.208 & 1.002 & \textcolor{blue}{\underline{0.436}} & \textcolor{blue}{\underline{0.560}} & 0.537 & 0.533 & 1.311 & 1.041 & 1.287 & 1.034 & 0.999 & 0.895 & 1.756 & 1.218 & 1.377 & 1.023 & 0.677 & 0.649 & \textbf{\textcolor{red}{0.110}} & \textbf{\textcolor{red}{0.165}} \\
                        & 48 & 1.741 & 1.247 & 2.001 & 1.257 & 1.699 & 1.146 & 1.733 & 1.216 & \textcolor{blue}{\underline{0.710}} & \textcolor{blue}{\underline{0.643}} & 0.731 & 0.655 & 1.610 & 1.169 & 1.651 & 1.185 & 1.483 & 1.078 & 2.620 & 1.609 & 2.703 & 1.591 & 0.919 & 0.813 & \textbf{\textcolor{red}{0.215}} & \textbf{\textcolor{red}{0.297}} \\
\hline
\multirow{3}{*}{\rotatebox{90}{ECL}}   & 1 & 2.911 & 1.617 & 4.279 & 1.968 & 1.897 & 1.232 & 2.708 & 1.549 & \textcolor{blue}{\underline{1.096}} & \textcolor{blue}{\underline{0.846}} & 1.101 & 0.849 & 6.354 & 2.371 & 6.384 & 2.327 & 6.045 & 2.355 & 311 & 16.18 & 29.88 & 4.99 & 4.639 & 2.141 & \textbf{\textcolor{red}{0.160}} & \textbf{\textcolor{red}{0.220}} \\
                        & 24 & 13.21 & 3.456 & 15.61 & 3.751 & 4.012 & 1.803 & 7.209 & 2.583 & 1.231 & \textcolor{blue}{\underline{0.809}} & \textcolor{blue}{\underline{1.211}} & 0.901 & 6.711 & 2.539 & 6.813 & 2.211 & 6.317 & 2.413 & 428 & 19.68 & 83.27 & 9.11 & 4.551 & 2.072 & \textbf{\textcolor{red}{0.263}} & \textbf{\textcolor{red}{0.264}} \\
                        & 48 & 25.98 & 4.906 & 15.88 & 3.849 & 4.877 & 2.008 & 9.243 & 2.840 & 1.600 & 1.065 & \textcolor{blue}{\underline{1.554}} & \textcolor{blue}{\underline{1.047}} & 7.342 & 2.596 & 7.523 & 2.474 & 7.321 & 2.557 & 469 & 19.65 & 144.89 & 11.43 & 5.819 & 2.355 & \textbf{\textcolor{red}{0.344}} & \textbf{\textcolor{red}{0.317}} \\
\hline
\multirow{3}{*}{\rotatebox{90}{Traffic}} & 1 & 0.298 & 0.489 & 0.279 & 0.498 & \textcolor{blue}{\underline{0.239}} & 0.458 & 0.280 & 0.509 & 0.580 & 0.562 & 0.662 & 0.614 & 0.602 & 0.676 & 0.653 & 0.698 & 0.511 & 0.615 & 0.612 & 0.712 & 0.259 & 0.381 & 0.273 & \textcolor{blue}{\underline{0.371}} & \textbf{\textcolor{red}{0.211}} & \textbf{\textcolor{red}{0.259}} \\
                        & 24 & 0.656 & 0.704 & 0.591 & 0.668 & 0.458 & 0.646 & 0.661 & 0.700 & 0.636 & 0.752 & 0.667 & 0.796 & 0.697 & 0.735 & 0.711 & 0.803 & 0.636 & 0.747 & 0.759 & 0.811 & 0.581 & 0.589 & \textcolor{blue}{\underline{0.369}} & \textcolor{blue}{\underline{0.386}} & \textbf{\textcolor{red}{0.307}} & \textbf{\textcolor{red}{0.281}} \\
                        & 48 & 0.791 & 0.868 & 0.633 & 0.756 & 0.517 & 0.690 & 0.759 & 0.828 & 0.800 & 0.868 & 0.732 & 0.856 & 0.724 & 0.801 & 0.783 & 0.824 & 0.684 & 0.774 & 0.814 & 0.831 & 0.701 & 0.633 & \textcolor{blue}{\underline{0.381}} & \textbf{\textcolor{red}{0.389}} & \textbf{\textcolor{red}{0.357}} & \textcolor{blue}{\underline{0.391}} \\
\hline
\multirow{3}{*}{\rotatebox{90}{Exchange}} & 1 & 0.579 & 0.561 & 0.887 & 0.842 & 0.836 & 0.814 & 0.432 & 0.557 & 0.502 & 0.639 & 0.531 & 0.649 & 0.847 & 0.820 & 0.859 & 0.827 & 0.778 & 0.782 & 0.460 & 0.578 & 0.275 & 0.424 & \textcolor{blue}{\underline{0.243}} & \textcolor{blue}{\underline{0.403}} & \textbf{\textcolor{red}{0.153}} & \textbf{\textcolor{red}{0.256}} \\
                        & 24 & 1.427 & 0.995 & 1.273 & 1.028 & 1.047 & 0.923 & \textcolor{blue}{\underline{0.980}} & 0.890 & 1.131 & 0.963 & 1.074 & 0.936 & 1.283 & 1.043 & 1.332 & 1.054 & 1.295 & 1.038 & 1.250 & 0.918 & 1.129 & \textcolor{blue}{\underline{0.863}} & 1.039 & 0.939 & \textbf{\textcolor{red}{0.803}} & \textbf{\textcolor{red}{0.618}} \\
                        & 48 & 2.473 & 1.273 & 2.711 & 1.547 & 2.279 & 1.410 & 1.748 & 1.122 & 1.693 & 1.201 & 1.576 & 1.155 & 2.371 & 1.440 & 2.162 & 1.370 & 1.884 & 1.273 & 2.379 & 1.242 & 1.507 & \textcolor{blue}{\underline{1.028}} & \textcolor{blue}{\underline{1.374}} & 1.072 & \textbf{\textcolor{red}{1.100}} & \textbf{\textcolor{red}{0.843}} \\
\bottomrule
\vspace{-1ex}
\end{tabular}}
\label{tab: main_table}
\vspace{-2ex}
\end{table*}

\begin{table*}[h]
\caption{Ablation studies of each component of \proposed (MSE).}
\vspace{-1ex}
\centering
\renewcommand{\arraystretch}{1}
    \resizebox{0.85\linewidth}{!}{
\begin{tabular}{ccc|ccc|ccc|ccc|ccc|ccc}
\toprule
 &&& \multicolumn{3}{c|}{ETTh2} & \multicolumn{3}{c|}{ETTm1} & \multicolumn{3}{c|}{WTH} & \multicolumn{3}{c|}{ECL} & \multicolumn{3}{c}{Traffic} \\
\cline{4-18}
 &$\mathcal{P}$& $\mathcal{T}$ & 1 & 24 & 48 & 1 & 24 & 48 & 1 & 24 & 48 & 1 & 24 & 48 & 1 & 24 & 48 \\
\hline
(1) & \textcolor{red}{\XSolidBrush}&\textcolor{red}{\XSolidBrush} & 0.927 & 3.121 & 4.790 & 0.211 & 0.561 & 0.666 & 0.238 & 0.595 & 0.767 & 3.467 & 3.479 & 3.688 & 0.524 & 0.767 & 0.804 \\
\hline
(2) & \textcolor{blue}{\Checkmark} & \textcolor{red}{\XSolidBrush} & 0.499 & 1.723 & 2.031 & 0.096 & 0.388 & 0.501 & 0.103 & 0.227 & 0.295 & 0.991 & 1.321 & 1.389 & 0.276 & 0.392 & 0.399 \\
\hline
(3) & \textcolor{red}{\XSolidBrush} & \textcolor{blue}{\Checkmark} & 0.718 & 2.321 & 2.889 & 0.122 & 0.481 & 0.544 & 0.179 & 0.431 & 0.601 & 2.131 & 2.773 & 2.908 & 0.411 & 0.603 & 0.640 \\
\hline
(4)-1 & \textcolor{blue}{\Checkmark}&\textcolor{blue}{\Checkmark} & 0.398 & \textbf{0.932} & \textbf{1.245} & \textbf{0.083} & \textbf{0.372} & \textbf{0.451} & 0.052 & \textbf{0.110} & \textbf{0.215} & \textbf{0.160} & \textbf{0.263} & \textbf{0.344} & 0.211 & \textbf{0.307} & \textbf{0.357} \\
\hline
(4)-2 & \textcolor{blue}{\Checkmark}&\textcolor{blue}{\Checkmark (\small STFT)} & \textbf{0.381} & 1.241 & 1.440 & 0.088 & 0.375 & 0.472 & \textbf{0.050} & 0.136 & 0.249 & 0.218 & 0.337 & 0.472 & \textbf{0.208} & 0.336 & 0.379 \\

\bottomrule
\end{tabular}}
\label{tab: ablation}
\vspace{-1ex}
\end{table*}

\subsection{Overall Performance}
The experimental results on eight datasets are summarized in Table~\ref{tab: main_table}. The reported results represent averages over three runs. We make the following key observations: \textbf{(1)} In settings without information leakage
, FSNet and OneNet, which are designed for OTSF, underperform static models such as DLinear, iTransformer, and TimeMixer.
This implies that FSNet and OneNet mainly exploit the leaked data on which they are trained—leading to rapid convergence—rather than truly acquiring the underlying structure of newly arriving patterns or learning how to adapt to them.
\textbf{(2)} Time series foundation models, i.e., Chronos-2 and TimesFM, underperform \proposed. Although these models exhibit strong zero-shot forecasting ability from large-scale pre-training, they lack an explicit mechanism to adapt to evolving target distributions in online scenarios. In contrast, \proposed~combines compositional spectral prompts with text descriptions, providing both distribution-aware guidance and recent contextual information for more effective adaptation.
\textbf{(3)} LLM-based models, i.e., LLM4TS, GPT4TS, and Time-LLM, substantially underperform \proposed. This gap arises because their alignment modules cannot handle continuous distribution shifts, leading to a breakdown in modality alignment. Conversely, \proposed~successfully adapts to these shifts using compositional spectral prompts and text descriptions, thereby preserving the crucial alignment between the LLM and the time series backbone.
\textbf{(4)} While the state-of-the-art OTSF method, DSOF, demonstrates strong performance compared with other baselines, it significantly underperforms \proposed. This demonstrates that while the dual-stream framework of DSOF prevents update delays, allowing the model to adapt quickly without information leakage, its emphasis on rapid convergence to incoming data limits its ability to capture underlying patterns,
leaving the model poorly equipped to adapt when data are scarce.
\textbf{(5)} \proposed~demonstrates robust performance across multiple datasets by combining the adaptability of a pre-trained LLM with compositional spectral prompts, which provide explicit guidance on the underlying time-series patterns under continuous distribution shifts.


\begin{table*}[t]
\caption{Comparison of MSE and MAE results across various datasets under scenarios with an extended online phase, where the train/valid/test split is set to 10\%/5\%/85\%.} 
\vspace{-1ex}
\centering
\renewcommand{\arraystretch}{1}
    \resizebox{0.95\linewidth}{!}{
\begin{tabular}{ccccc|ccc|ccc|ccc|ccc}
\toprule
 && \multicolumn{3}{c|}{ETTh2} & \multicolumn{3}{c|}{ETTm1} & \multicolumn{3}{c|}{WTH} & \multicolumn{3}{c|}{ECL} & \multicolumn{3}{c}{Traffic} \\
\cline{3-17}
 & & 1 & 24 & 48 & 1 & 24 & 48 & 1 & 24 & 48 & 1 & 24 & 48 & 1 & 24 & 48 \\
\hline
\multirow{2}{*}{FSNet} & MSE & 15.231 & 24.591 & 28.935 & 0.399 & 2.452 & 5.113 & 1.221 & 2.341 & 4.512 & 299.34 & 435.12 & 458.98 & 1.212 & 1.751 & 1.999 \\
& MAE & 3.802 & 4.758 & 5.179 & 0.516 & 1.365 & 2.061 & 1.004 & 1.330 & 2.041 & 15.301 & 17.859 & 20.423 & 1.001 & 1.232 & 1.353 \\
\hline
\multirow{2}{*}{OneNet} & MSE & 4.240 & 9.342 & 13.582 & 0.278 & 1.428 & 1.677 & 0.813 & 1.535 & 2.991 & 30.091 & 82.29 & 142.273 & 0.531 & 0.778 & 0.933 \\
& MAE & 2.009 & 3.006 & 3.485 & 0.427 & 1.074 & 1.204 & 0.801 & 1.098 & 1.599 & 5.285 & 8.041 & 10.797 & 0.588 & 0.742 & 0.905 \\
\hline
\multirow{2}{*}{DSOF} & MSE & 0.851 & 2.889 & 5.028 & 0.173 & 0.561 & 0.779 & 0.313 & 0.711 & 1.137 & 4.898 & 5.173 & 6.092 & 0.499 & 0.711 & 0.793 \\
& MAE & 0.872 & 1.499 & 2.042 & 0.397 & 0.668 & 0.846 & 0.539 & 0.803 & 1.006 & 2.134 & 2.204 & 2.368 & 0.656 & 0.759 & 0.801 \\
\hline
\multirow{2}{*}{\proposed} & MSE & \textbf{0.412} & \textbf{0.978} & \textbf{1.305} & \textbf{0.087} & \textbf{0.411} & \textbf{0.487} & \textbf{0.051} & \textbf{0.103} & \textbf{0.223} & \textbf{0.179} & \textbf{0.291} & \textbf{0.362} & \textbf{0.485} & \textbf{0.356} & \textbf{0.379} \\
& MAE & \textbf{0.434} & \textbf{0.613} & \textbf{0.730} & \textbf{0.194} & \textbf{0.601} & \textbf{0.637} & \textbf{0.206} & \textbf{0.291} & \textbf{0.452} & \textbf{0.403} & \textbf{0.509} & \textbf{0.581} & \textbf{0.434} & \textbf{0.481} & \textbf{0.578} \\
\bottomrule
\end{tabular}}
\label{tab: few_shot}
\vspace{-2ex}
\end{table*}

\subsection{Ablation Study}
\label{sec: ablation}
To assess the impact of the compositional spectral prompt (i.e., $\mathcal{P}$) and text description (i.e., $\mathcal{T}$) in \proposed, Table~\ref{tab: ablation} presents ablation studies across five cases, including the vanilla \proposed~(Row (4)-1), with key observations as follows:
\textbf{(1)} Introducing the compositional spectral prompt is helpful (Row (1) vs. (2)). 
Given the sequential nature of time-series data, in which continuous distribution shifts are inevitable, providing the model with compositional spectral prompts that capture these dynamics is highly effective. 
Specifically, 
\proposed~captures overall time-series patterns by learning spectral basis prompts grounded in decomposed frequency bases.
This enables effective adaptation to unseen patterns by compositionally recombining the learned spectral basis prompts,
resulting in nearly a 50\% improvement in MSE. We further analyze the advantage of frequency-domain prompting over time-domain prompting in Section~\ref{sec: time_prompting}. 
\textbf{(2)} Providing recent pattern information to the LLM in the form of text descriptions is effective (Row (1) vs. (3)). The text description serves as a practical auxiliary modality for online forecasting, as it injects dynamically changing contextual information into the LLM beyond raw numerical observations, enriching data-scarce online scenarios without additional training and enabling effective adaptation.
\textbf{(3)} Leveraging both compositional spectral prompts and text description together can yield synergistic effects (Row (2\&3) vs. (4)-1). Through compositional spectral prompts, the overall pattern of the given time series is captured, while text description provides information on recent patterns in the time and frequency domains, enabling the model to effectively adapt to recent patterns without being hindered by distribution shifts. \textbf{(4)} Utilizing DWT instead of STFT is more effective for providing recent pattern information from a frequency perspective (Row (4)-1 vs. (4)-2). STFT struggles with temporal resolution due to its use of a fixed window size, whereas DWT adapts the window size based on the frequency of the time series, making it more effective in capturing non-stationary signals. This results in comparable performance in relatively easy tasks with a prediction horizon of 1. However, for more challenging tasks with longer horizons, where capturing the underlying recent patterns is crucial, the use of DWT proves to be more effective. 


\begin{table}[t]
\caption{Comparison of MSE and MAE results in a cross-dataset scenario, where different datasets from the same ETT domain are used in the training and online phases to induce distribution shifts, with the prediction horizon (i.e., $H$) set to 1. $Training$ refers to the dataset used in the training phase, while $Online$ refers to the dataset used in the online phase.}
\vspace{-1ex}
\centering
\renewcommand{\arraystretch}{1.2} 
\resizebox{0.99\linewidth}{!}{
\begin{tabular}{lc||cc|cc|cc|cc}
\toprule
\multirow{2}{*}{\textit{Training}} & \multirow{2}{*}{\textit{Online}} & \multicolumn{2}{c|}{FSNet} & \multicolumn{2}{c|}{OneNet} & \multicolumn{2}{c|}{DSOF} & \multicolumn{2}{c}{\proposed} \\
\cline{3-10}
& & MSE & MAE & MSE & MAE & MSE & MAE & MSE & MAE \\
\hline
\hline
\multirow{2}{*}{\textbf{ETTh1}} & $\rightarrow$ ETTh2 & 13.664 & 3.124 & 4.025 & 1.862 & 0.993 & 0.806 & \textbf{0.427} & \textbf{0.453} \\
& $\rightarrow$ ETTm2 & 1.815 & 1.132 & 1.732 & 1.016 & 0.723 & 0.702 & \textbf{0.178} & \textbf{0.319} \\
\hline
\multirow{2}{*}{\textbf{ETTh2}} & $\rightarrow$ ETTh1 & 12.667 & 3.011 & 3.833 & 1.808 & 0.901 & 0.849 & \textbf{0.399} & \textbf{0.531} \\
& $\rightarrow$ ETTm2 & 1.791 & 1.029 & 1.489 & 1.022 & 0.818 & 0.744 & \textbf{0.211} & \textbf{0.359} \\
\hline
\multirow{2}{*}{\textbf{ETTm1}} & $\rightarrow$ ETTh2 & 15.371 & 3.571 & 3.989 & 1.610 & 1.003 & 0.901 & \textbf{0.521} & \textbf{0.621} \\
& $\rightarrow$ ETTm2 & 1.335 & 1.003 & 0.989 & 0.861 & 0.542 & 0.536 & \textbf{0.128} & \textbf{0.257} \\
\hline
\multirow{2}{*}{\textbf{ETTm2}} & $\rightarrow$ ETTh2 & 14.989 & 3.199 & 4.138 & 1.734 & 1.211 & 0.945 & \textbf{0.513} & \textbf{0.616} \\
& $\rightarrow$ ETTm1 & 1.299 & 1.039 & 1.315 & 0.946 & 0.517 & 0.619 & \textbf{0.132} & \textbf{0.263} \\
\bottomrule
\vspace{-2ex}
\end{tabular}}
\label{tab: zero_shot}
\vspace{-3ex}
\end{table}

\subsection{Further Analysis}


\smallskip
\subsubsection{Robustness to the extension of the online phase.}
\label{sec: extended_online}
In Table~\ref{tab: few_shot}, we analyze the model's robustness when the online phase is extended. Specifically, we evaluate performance by adjusting the OTSF train/valid/test split from 20\%/5\%/75\% to 10\%/5\%/85\%, extending the online phase.
Prior methods (i.e., FSNet, OneNet, and DSOF) experience significant performance degradation compared to the results in Table~\ref{tab: main_table} for the following two reasons: (1) They are incapable of storing all the recurring patterns in an associative memory as the online phase is extended, and (2) the continuous occurrence of unseen patterns prevents the model from maintaining adaptability. In contrast, \proposed, thanks to the rich knowledge and transferability of the pre-trained LLM, effectively maintains adaptability even in data-scarce online scenarios, demonstrating performance comparable to the results in Table~\ref{tab: main_table}. 
We argue that adaptability to an extended online phase is enhanced by two key designs: 
(i) compositional spectral prompting, which represents evolving patterns using a finite set of spectral basis prompts,  
and (ii) a text description of recent patterns provided to the LLM, which compensates for data scarcity and enables efficient adaptation without additional training.

\subsubsection{Robustness to distribution shifts.}
\label{sec: cross_dataset}
Table~\ref{tab: zero_shot} shows the cross-dataset experiments using the ETT datasets. 
ETTh1 and ETTh2 are hourly measurements from two different electricity transformers, while ETTm1 and ETTm2 are their corresponding 15-minute-resolution variants. Therefore, each pair shares the same sampling resolution but exhibits different temporal patterns due to differences in the underlying transformers.
To induce distribution shifts, we deliberately use different datasets for the training and online phases. The model learns the base knowledge from the training data and adapts to the streaming online data, with the prediction horizon set to 1. 
We observe that \proposed~outperforms all baselines across the 8 scenarios. FSNet, OneNet, and DSOF rely on associative memory to store recurring patterns and adapt by retrieving similar ones. This approach fails when unseen patterns arise, as no meaningful associations can be found, leading to substantial performance degradation (compare with the results in Table~\ref{tab: main_table}). In contrast, \proposed~represents patterns through compositional spectral prompts by learning a spectral basis prompt for each frequency basis, rather than directly storing the time series patterns. As a result, even unseen patterns can be reliably expressed as new compositions of spectral basis prompts, making the model robust to distribution shifts and allowing its performance to remain largely consistent with the results in Table~\ref{tab: main_table}.


\subsubsection{Online Updating Cost.}
\label{sec: online_update}
To analyze the efficiency of \proposed, we compare the number of updated parameters during the online phase and runtime statistics with those of existing online time series forecasting models.

\begin{table}[h]
\small
\caption{The number of parameters updated during the online phase of OTSF models for prediction horizon 1 on the ETTh2 dataset.}
\vspace{-2ex}
\centering
\renewcommand{\arraystretch}{1.1}
    \resizebox{0.65\linewidth}{!}{
\begin{tabular}{cccc}
\toprule
  FSNet & OneNet & DSOF & \proposed \\
\hline
  2,037,115 & 1,018,045 & 1,236,349 & 897 \\
\bottomrule
\end{tabular}}
\label{tab: training_parameter}
\end{table}

\vspace{-1ex}
\paragraph{Parameters.}
Table~\ref{tab: training_parameter} reports the number of parameters updated during the online phase (i.e., those that are not frozen) for each OTSF model in experiments using the ETTh2 dataset. \proposed~updates significantly fewer parameters compared to the baselines, indicating its efficiency. This demonstrates that \proposed~can leverage model adaptability to achieve strong performance (see Table~\ref{tab: main_table}) with fewer updated parameters. Updating many parameters during the online phase can make models vulnerable to distribution shifts and cause them to forget the base knowledge learned during training phase. 
In this regard, \proposed~effectively and efficiently adapts to new data without forgetting previously learned knowledge by updating only a small subset of parameters.

\begin{table}[h]
\vspace{-1ex}
\caption{Comparison of runtime statistics between \proposed~and existing OTSF methods for the scenario with a prediction horizon of 1 on each dataset, utilizing the total Training Phase Duration (sec), the total Online Phase Duration (sec), and the Inference Latency (sec/itr), defined as the time required per update.}
\vspace{-2ex}
\centering
\renewcommand{\arraystretch}{1.1}
    \resizebox{0.99\linewidth}{!}{
\begin{tabular}{ccccccccc}
\toprule
 & Metric & ETTh1 & ETTh2 & ETTm1 & ETTm2 & WTH & ECL & Traffic \\
\hline
\multirow{4}{*}{\rotatebox{90}{FSNet}} & Training Phase Duration (sec) & 275 & 274 & 801 & 830 & 775 & 398 & 159 \\
                        & Online Phase Duration (sec) & 341 & 332 & 1,024 & 1,019 & 998 & 463 & 351 \\
                        & Inference Latency (sec/itr) & 0.031 & 0.030 & 0.022 & 0.022 & 0.025 & 0.023 & 0.026 \\
\hline
\multirow{4}{*}{\rotatebox{90}{OneNet}} & Training Phase Duration (sec) & 551 & 540 & 1,621 & 1,633 & 1,503 & 848 & 303 \\
                        & Online Phase Duration (sec) & 701 & 690 & 2,237 & 2,241 & 2,079 & 994 & 741 \\
                        & Inference Latency (sec/itr) & 0.063 & 0.061 & 0.050 & 0.050 & 0.051 & 0.050 & 0.053 \\
                        \hline
\multirow{4}{*}{\rotatebox{90}{DSOF}} & Training Phase Duration (sec) & 622 & 613 & 1,904 & 1,936 & 1,789 & 994 & 379 \\
                        & Online Phase Duration (sec) & 733 & 705 & 2,588 & 2,559 & 2,371 & 1,201 & 855 \\
                        & Inference Latency (sec/itr) & 0.064 & 0.062 & 0.058 & 0.056 & 0.057 & 0.059 & 0.062 \\
\hline
\multirow{4}{*}{\rotatebox{90}{\proposed}} & Training Phase Duration (sec) & 4,650 & 4,650 & 14,428 & 14,501 & 13,786 & 7,249 & 2,788 \\
                        & Online Phase Duration (sec) & 787 & 741 & 2,938 & 2,973 & 2,841 & 1,389 & 931 \\
                        & Inference Latency (sec/itr) & 0.068 & 0.065 & 0.064 & 0.066 & 0.068 & 0.065 & 0.067 \\
\bottomrule
\end{tabular}}
\label{tab: runtime_stat}
\vspace{-2ex}
\end{table}


\paragraph{Runtime Comparison.}
Table~\ref{tab: runtime_stat} presents the runtime statistics for each dataset in the scenario where the prediction horizon is 1. We utilize the total Training Phase Duration (sec), the total Online Phase Duration (sec), and the Inference Latency (sec/itr), defined as the time required per update. OTSF requires a model to be sufficiently pre-trained on initial data (i.e., training phase) and then continually adapted to the subsequent data stream (i.e., online phase). Consequently, resource optimization must prioritize the cost-sensitive online phase over the training phase. The duration of the online phase and the inference latency for \proposed~require only slightly more time compared to existing OTSF models. This demonstrates that \proposed~continuously adapts to new online distributions with significantly less parameter tuning (see Table~\ref{tab: training_parameter}) by simultaneously leveraging the LLM's superior transferability and the distribution guidance provided by compositional spectral prompts. While \proposed~incurs a higher initial cost for training duration, this cost is justifiable as the training phase is less resource-sensitive and prioritizes the sufficient acquisition of base knowledge.
In summary, \proposed~achieves significantly superior performance (please refer to Table~\ref{tab: main_table}) in the cost-sensitive online phase with only a comparable cost, despite its larger initial overhead during the initial training.

\begin{table}[h]
\small
\caption{Average cosine similarity between prompts in the prompt bank, generated by three strategies, for the prediction horizon1 in the ETTh2 dataset.}
\vspace{-1ex}
\centering
\renewcommand{\arraystretch}{1.1}
    \resizebox{0.6\linewidth}{!}{
\begin{tabular}{ccc}
\toprule
   & $\mathbf{P}^2_{low}$ & $\mathbf{P}^3_{low}$ \\
\hline
  Similarity with $\mathbf{P}^1_{low}$ & 0.7375 & -0.1827 \\
\bottomrule
\end{tabular}}
\label{tab: prompt_exp}
\vspace{-2ex}
\end{table}

\subsubsection{Generalizability of Compositional Spectral Prompting.}
\label{sec: generalizability}
In Table~\ref{tab: prompt_exp}, to demonstrate that the composition of spectral basis prompts can effectively represent unseen patterns
in the online phase, even with learning solely from the training phase, we compare three Spectral Prompt Banks (i.e., $\mathbf{P}^1_{low}$, $\mathbf{P}^2_{low}$, and $\mathbf{P}^3_{low}$) generated using the following strategies:

\begin{itemize}[leftmargin=10pt]
\item $\mathbf{P}^1_{low}$ (Optimal): The prompt bank is trained during the training phase and further trained during the online phase.
\item $\mathbf{P}^2_{low}$ (\proposed): The prompt bank is trained during the training phase, and frozen during the online phase.
\item $\mathbf{P}^3_{low}$: The prompt bank is randomly generated.
\end{itemize}

\noindent The cosine similarity between the optimal Prompt Bank 1 (i.e., $\mathbf{P}^1_{low}$) and each of Prompt Bank 2 (i.e., $\mathbf{P}^2_{low}$) and Prompt Bank 3 (i.e., $\mathbf{P}^3_{low}$) are shown in Table~\ref{tab: prompt_exp}. $\mathbf{P}^1_{low}$ is the optimal prompt bank that can be obtained when unseen patterns (i.e., patterns in online phase) are included in the training. $\mathbf{P}^2_{low}$ shows a high degree of similarity to the $\mathbf{P}^1_{low}$, both in absolute terms and especially when compared to $\mathbf{P}^3_{low}$. 
This indicates that, even without additional learning during the online phase, compositional spectral prompting can effectively represent incoming new patterns by recombining knowledge learned from frequency bases.

\subsubsection{Comparison of frequency- and time-domain prompting.}
\label{sec: time_prompting}
To demonstrate that compositional spectral prompting effectively captures underlying time-series patterns, we compare its frequency-domain construction with an alternative time-domain prompting strategy. 
The time-domain strategy replaces frequency bases with representation-space prototypes derived from the time-series backbone. 
During training, input sequences are encoded by the pre-trained time-series backbone and clustered in the representation space, where each cluster centroid is treated as a distributional prototype and assigned a learnable prompt. 
During the online phase, each incoming instance is encoded into the same representation space, matched to the nearest prototype, and guided by the corresponding prompt. 
Thus, the time-domain strategy retrieves a prompt based on representation similarity, whereas our frequency-domain construction composes spectral basis prompts according to the amplitudes of decomposed frequency components.
The frequency- and time-domain prompts are trained under two distinct settings:

\begin{itemize}[leftmargin=10pt]
\item $S_1$ (Optimal): The prompt bank is trained during the training phase and further trained during the online phase.
\item $S_2$ (Freezing): The prompt bank is trained during the training phase, and frozen during the online phase.
\end{itemize}

Based on these settings, we define four variations of prompt banks as follows:

\begin{itemize}[leftmargin=15pt]
    \item $\mathbf{P}^{S_1}_{freq}$ and $\mathbf{P}^{S_2}_{freq}$: Frequency basis-driven prompt banks generated under the $S_1$ and $S_2$ settings, respectively.
    \item $\mathbf{P}^{S_1}_{time}$ and $\mathbf{P}^{S_2}_{time}$: Cluster-based time-domain prompt banks generated under the $S_1$ and $S_2$ settings, respectively.
\end{itemize}

\begin{table}[t]
\centering
\caption{Performance and the total time (in seconds) required for the online phase of frequency- and time-domain prompting.}
\label{tab: time_domain_prompting}
\vspace{-1.5ex}
\renewcommand{\arraystretch}{1.1}
\setlength{\tabcolsep}{3pt} 
\resizebox{0.95\linewidth}{!}{ 
\begin{tabular}{l ccc ccc ccc ccc}
\toprule
 & \multicolumn{3}{c}{$\mathbf{P}^{S_1}_{freq}$} & \multicolumn{3}{c}{$\mathbf{P}^{S_2}_{freq}$} & \multicolumn{3}{c}{$\mathbf{P}^{S_1}_{time}$} & \multicolumn{3}{c}{$\mathbf{P}^{S_2}_{time}$} \\
\cmidrule(lr){2-4} \cmidrule(lr){5-7} \cmidrule(lr){8-10} \cmidrule(lr){11-13}
Metric & 1 & 24 & 48 & 1 & 24 & 48 & 1 & 24 & 48 & 1 & 24 & 48 \\
\midrule
MSE & 0.372 & 0.859 & 1.148 & 0.398 & 0.932 & 1.245 & 0.395 & 1.101 & 1.263 & 0.635 & 1.597 & 1.924 \\
MAE & 0.402 & 0.582 & 0.659 & 0.423 & 0.607 & 0.699 & 0.598 & 1.001 & 0.923 & 0.756 & 1.076 & 1.273 \\
\midrule
Time & 936 & 1,028 & 1,330 & 741 & 811 & 837 & 1,031 & 1,141 & 1,438 & 843 & 931 & 958 \\
\bottomrule
\end{tabular}
}
\vspace{-1ex}
\end{table}

\begin{table}[t]
\small
\caption{Cosine similarity between finalized prompts of $S_1$ and $S_2$ across frequency and time domains.}
\vspace{-1ex}
\centering
\renewcommand{\arraystretch}{1.1}
    \resizebox{0.75\linewidth}{!}{
\begin{tabular}{ccc}
\toprule
   & $\textup{Sim}(\mathbf{P}^{S_1}_{freq}, \mathbf{P}^{S_2}_{freq})$ & $\textup{Sim}(\mathbf{P}^{S_1}_{time}, \mathbf{P}^{S_2}_{time})$ \\
\hline
  Cosine Similarity & 0.7375 & 0.2481 \\
\bottomrule
\end{tabular}}
\label{tab: prompt_sim}
\vspace{-1ex}
\end{table}

\noindent Table~\ref{tab: time_domain_prompting} presents a comparative analysis of the performance of four variants and their cumulative online phase duration for prediction lengths \{1, 24, 48\} on the ETTh2 dataset.
Across both domains, continuously updating prompts during the online phase (i.e., $S_1$) yields superior performance, albeit at the cost of reduced efficiency. In contrast, when prompts are learned only during the training phase and kept frozen during online inference (i.e., $S_2$), the time-domain prompting strategy exhibits a substantial performance degradation. This indicates that the time-domain approach, which relies on clustering representations to characterize data distributions, fails to yield a representative prompt for the data distribution when shifts occur. Conversely, frequency-domain prompting remains robust, as it represents emerging distributions as combinations of underlying frequency bases, thereby preserving performance even in the presence of distributional shifts. Additionally, we confirmed this difference from the perspective of the learned prompt itself. In table~\ref{tab: prompt_sim}, we compared the similarity between the prompt finalized after the online phase was completed in $S_1$ and the prompt finalized after the end of the training phase in $S_2$, across both domains. Compared to the time domain, the frequency domain prompt exhibits a significantly higher similarity between the learned prompts in both cases. This demonstrates that the effective learning of the underlying basis during the training phase alone enables the model to adequately handle the shift without additional learning in the online phase.

\subsubsection{Sensitivity Analysis.}
\label{sec: sensitivity_analysis}
We present a sensitivity analysis for the hyperparameters $\gamma$ (in Equation~\ref{eq: pattern_embedding}) and $\delta$ (in Equation~\ref{eq: online_loss}) utilized in \proposed.

\paragraph{Hyperparameter $\gamma$}

\begin{figure}
  \centering
  \includegraphics[width=0.99\linewidth]{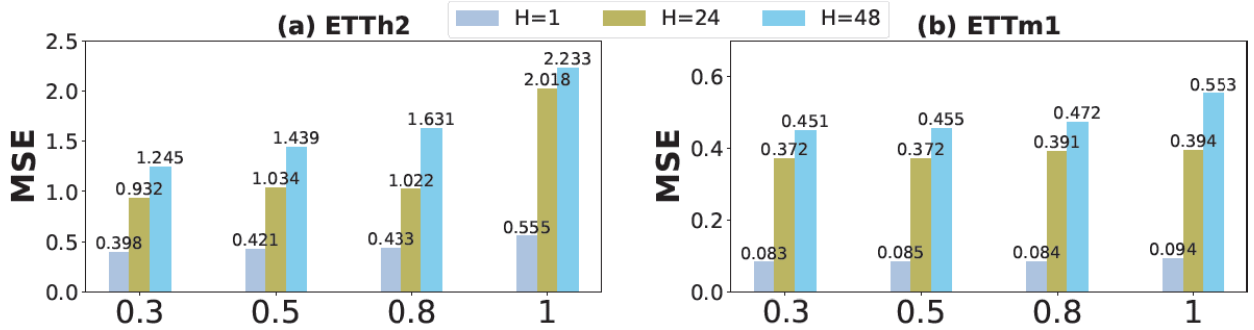}
  \vspace{-1ex}
  \caption{Sensitivity analysis of $\gamma$ in Equation~\ref{eq: pattern_embedding}. MSE results are reported on (a) ETTh2 and (b) ETTm1 for $H \in \{1, 24, 48\}$ and $\gamma \in \{0.3, 0.5, 0.8, 1\}$.}
  \label{fig: gamma}
\vspace{-3ex}
\end{figure}

To analyze the sensitivity of \proposed~to the hyperparameter $\gamma$, which is used to filter for significant low-frequency information when constructing compositional spectral prompts, we conduct an experiment presented in Figure~\ref{fig: gamma}. Using the ETTh2 and ETTm1 datasets, we vary the $\gamma$ parameter in Equation~\ref{eq: pattern_embedding} across values of 0.3, 0.5, 0.8, and 1, and observe the corresponding MSE for prediction horizons of 1, 24, and 48. On both datasets, optimal performance is achieved when $\gamma=0.3$. Performance degrades as more high-frequency components are retained, with the most significant drop observed when $\gamma=1$. This indicates that when constructing compositional spectral prompts from frequency components to capture the overall distribution of the time series data,
the high-frequency components are largely irrelevant to the overall distribution and instead act as noise.

\begin{figure}
  \centering
  \includegraphics[width=0.99\linewidth]{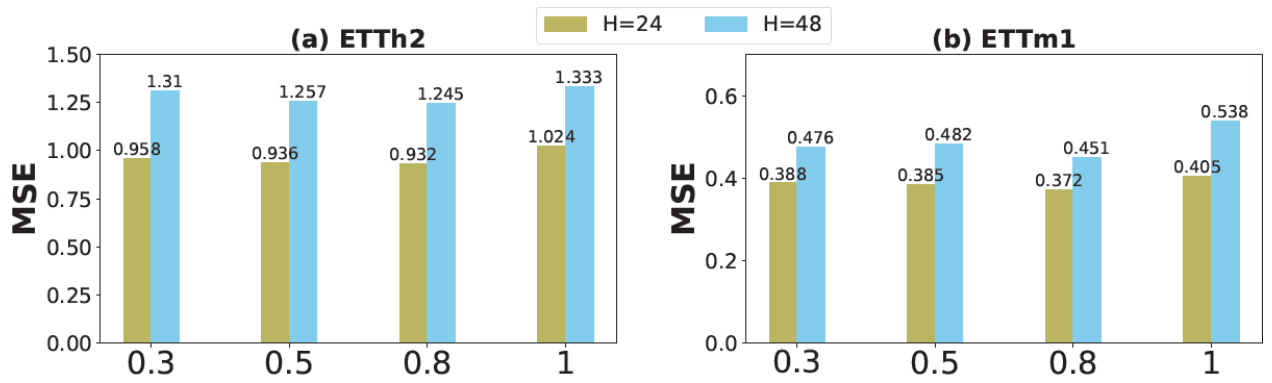}
  \vspace{-1ex}
  \caption{Sensitivity analysis of the geometric decay factor $\delta$ in Equation~\ref{eq: online_loss}. MSE results are reported on (a) ETTh2 and (b) ETTm1 for $H \in \{24, 48\}$ and $\delta \in \{0.3, 0.5, 0.8, 1\}$.}
  \label{fig: delta}
\vspace{-4ex}
\end{figure}

\paragraph{Geometric Decay Factor $\delta$}

To investigate the sensitivity of \proposed ~to the geometric decay factor $\delta$, which is utilized to reduce the influence of pseudo-label unreliability and prediction errors for timestamps distant from the current observation when the prediction horizon is greater than 1, we conduct an experiment presented in Figure~\ref{fig: delta}. Using the ETTh2 and ETTm1 datasets, we vary the $\delta$ parameter in Equation~\ref{eq: online_loss} across values of 0.3, 0.5, 0.8, and 1 and observe the corresponding MSE for prediction horizons of 24 and 48. On both datasets, optimal performance is achieved when $\delta$ is around 0.8. A $\delta$ value of 1 which signifies no decay effect, leads to a performance drop. This is due to the negative influence of unreliable pseudo-labels generated by the time series backbone and the less accurate predictions for distant timestamps. Therefore, to robustly adapt to continuous distribution shifts, we utilize an appropriate geometric decay factor to mitigate noise during the model's training process. We observed that a similar value of $\delta$ (i.e., $\delta=0.8$) is consistently effective across various datasets.

\section{Conclusion}

In this paper, we present the first LLM-based OTSF framework, called \proposed, which excels in both continuous distribution shifts and extended online scenarios. We devise a compositional spectral prompting strategy that captures the overall distribution (i.e., overall pattern) of each input by recombining learnable spectral basis prompts according to its frequency-domain structure. These distribution-aware prompts guide the pre-trained LLM to adapt its rich knowledge and transferability to continuously shifting online distributions, enabling efficient rapid adaptation in data-scarce scenarios. Furthermore, text descriptions containing recent pattern information enrich the limited online data without requiring additional training. \proposed~demonstrates promising performance across various real-world datasets and exhibits robust performance against distribution shifts and the extension of the online phase, highlighting its applicability to real-world online forecasting scenarios.

\begin{acks}
This work was supported by Institute of Information \& communications Technology Planning \& Evaluation (IITP) grant funded by the Korea government(MSIT) (RS-2022-II220157), National Research Foundation of Korea(NRF) grant funded by the Korea government(MSIT) (RS-2024-00406985), and National Research Foundation of Korea(NRF) funded by Ministry of Science and ICT (RS-2022-NR068758).
\end{acks}


\section*{GenAI Disclosure}
We acknowledge the limited use of LLMs (e.g., GPT-5 and Claude) for (1) improving the grammar, clarity, and stylistic variation of this paper, as well as reducing its length to comply with page limits, and (2) minor code refactoring and debugging for plotting and visualization. All AI-assisted revisions and code modifications were carefully reviewed and validated by the authors. The core ideas, methodology, experiments, and interpretations presented in this work are entirely original contributions of the authors.

\bibliographystyle{ACM-Reference-Format}
\balance
\bibliography{reference}

\end{document}